# PiClick: Picking the desired mask from multiple candidates in click-based interactive segmentation

Cilin Yan[a,1], Haochen Wang[b,1], Jie Liu[b], Xiaolong Jiang[c], Yao Hu[c], Xu Tang[c], Guoliang Kang[a,*] and Efstratios Gavves[b]

[a]*Beihang University, Beijing, China*
[b]*University of Amsterdam, Nederland*
[c]*Xiaohongshu, Beijing, China*

**ABSTRACT**

Click-based interactive segmentation aims to generate target masks via human clicking, which facilitates efficient pixel-level annotation and image editing. In such a task, target ambiguity remains a problem hindering the accuracy and efficiency of segmentation. That is, in scenes with rich context, one click may correspond to multiple potential targets, while most previous interactive segmentors only generate a single mask and fail to deal with target ambiguity. In this paper, we propose a novel interactive segmentation network named *PiClick*, to yield all potentially reasonable masks and suggest the most plausible one for the user. Specifically, PiClick utilizes a Transformer-based architecture to generate all potential target masks by mutually interactive mask queries. Moreover, a Target Reasoning module (TRM) is designed in PiClick to automatically suggest the user-desired mask from all candidates, relieving target ambiguity and extra-human efforts. Extensive experiments on 9 interactive segmentation datasets demonstrate PiClick performs favorably against previous state-of-the-arts considering the segmentation results. Moreover, we show that PiClick effectively reduces human efforts in annotating and picking the desired masks. To ease the usage and inspire future research, we release the source code of PiClick together with a plug-and-play annotation tool at https://github.com/cilinyan/PiClick.

## 1. Introduction

Interactive image segmentation [37, 25, 16, 52, 28] aims to obtain accurate binary segmentation masks based on user interactions, serving as a crucial tool for obtaining high-quality pixel-level annotations and enhancing visual perception tasks [47, 53]. There are various interaction types for interactive segmentation, including bounding boxes [54], polygons [1], clicks [45], scribbles [51, 49, 48], and combinations of these [56]. The click-based approach [37, 35] is most popular due to its simplicity and ease of use for specifying the desired object.

In click-based interactive segmentation, users initially click on the object they want to segment and optionally click on the background for negative indication. These inputs guide a segmentation algorithm, such as graph cuts [43], random walks [18], or deep learning models [50], to interpret the spatial context of the clicks relative to the image. Initial results often need refinement, allowing users to iteratively enhance segmentation by adding further positive or negative clicks. Each interaction with the model mainly consists of the image and all click positions [37, 9, 58, 44]. Some methods [37, 9] further incorporate the predicted mask from the previous interaction.

Recent efforts for click-based interactive segmentation mainly focus on obtaining detailed object masks. The FocusCut [35] conducts local refinement with focus views to improve the details of the segmentation masks. The

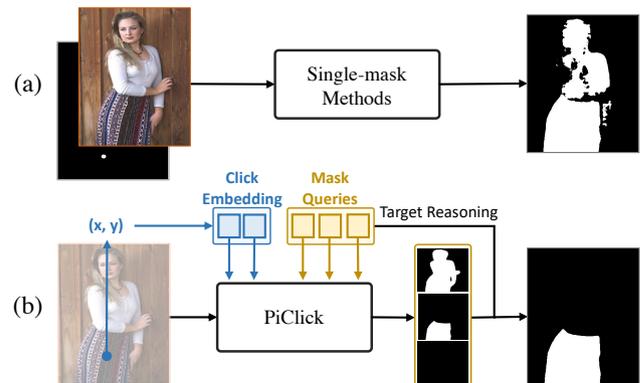

**Figure 1:** (a) Existing single-mask methods typically generate a single segmentation mask, which cannot mitigate the target ambiguity issue, e.g., clicks on the skirt may refer to the skirt or the person wearing the skirt. (b) Our PiClick obtains multiple semantic meaningful masks with mask queries, and then picks out the desired one as target by Target Reasoning module.

SimpleClick [37] captures global information with plain ViT [13] backbones for precise segmentation. Those methods typically output one single mask based on user clicks (thus known as the single-mask methods), thereby remaining vulnerable to target ambiguity. As demonstrated in Fig. 1 (a), target ambiguity denotes that in images containing rich semantic context, one user click usually refers to a set of potential segmentation masks instead of a single one, *e.g.*, a click on the skirt may refer to the skirt or the person wearing the skirt. Most previous methods predict a single mask, which are easily confused by the ambiguous clicks. They also ignore the fact that different click positions may indicate different intentions of users and thus usually fail to output a mask best matching the users' intention.

*Corresponding author. (e-mail: kgl.prml@gmail.com)
ORCID(s): 0000-0003-1978-2025 (G. Kang)
[1]Equal contribution.

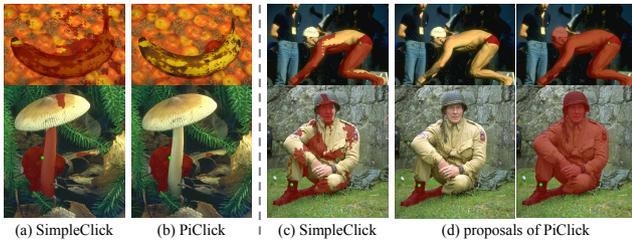

**Figure 2:** Segmentation mask generated by SimpleClick [37] and our PiClick.

Several previous works [31, 29] try to solve the target ambiguity problem by adopting separate convolutional branches to generate multiple segmentation masks. However, these convolutional methods lack the ability to generate diverse set predictions, as multiple convolutional branches are trained separately, which cannot exploit inter-mask interactions to collaboratively generate multiple masks. Consequently, convolutional methods tend to output similar masks and also require extra human interventions [31, 29] to select the desired mask. Recently, [50] utilizes the centrality of clicks to address the target ambiguity problem. However, they employ a two-stage approach rather than training in an end-to-end manner.

To effectively solve target ambiguity in interactive segmentation, we hereby introduce PiClick, a click-based Transformer architecture to generate diverse segmentation masks and then automatically select the best-matched one with a Target Reasoning module (TRM), see Fig. 1 (b). Specifically, PiClick encodes clicks as raw binary representation (binary disks with a small radius, referred to as disk maps) and feature-level representation (referred to as click position-aware embeddings). The former is fused with images via an image encoder, producing click-aware image features. The latter is used in the decoding process for improved localization precision. Then a set of learnable mask queries interact with the click-aware image feature and click position-aware embeddings within a Transformer decoder to propose diverse masks. Thanks to the mask queries, sophisticated designs for encouraging output diversity such as the diversity loss [29], multi-scale output branches [31], and non-maximum-suppression [31] can be eliminated, thus significantly simplifying the training and inference pipeline, as well as improving the applicability of interactive segmentation. Finally, TRM takes mask queries as input to predict the intersection over union (IoU) between each predicted mask and the potential target mask to select the best-matched one. We experimentally show that the results of TRM align well with those of human selection, proving PiClick can effectively capture user intention.

To comprehensively evaluate PiClick, we run experiments on 6 natural interactive segmentation benchmarks as well as 3 medical benchmarks. In these experiments, our method outperforms the state-of-the-art methods (SOTA) [37, 45, 9] on 7 benchmarks and performs comparably with SOTA methods on the other 2 benchmarks. To further ease the usage of PiClick and facilitate more efficient annotation, we also develop a plug-and-play annotation platform based on PiClick.

To sum up, our contributions are shown as follows:

(i) We propose a Transformer-based architecture named PiClick to generate multiple diverse masks to mitigate the target ambiguity issue.

(ii) A Target Reasoning module (TRM) is designed in PiClick to mimic or approach the human behavior of mask selection to reduce human efforts and improve the efficiency of mask annotations in practice.

(iii) PiClick performs favorably against previous state-of-the-arts on 9 interactive segmentation benchmarks. The code of PiClick and the corresponding annotation platform are publicly released.

## 2. Related work
### 2.1. Interactive Image Segmentation

Interactive image segmentation segments target objects based on user inputs. Early solutions typically exploit boundary properties for segmentation [24, 27] or adopt graphical models, such as graph cut [5], random walker [18], and geodesic approaches [19]. These approaches rely on low-level features and usually result in poor segmentation quality, especially in complex scenarios. Recently, integrating deep learning into interactive segmentation [30, 31, 36] has brought promising improvements. Concretely, the method proposed in [55] first adopts a CNN-based model for interactive segmentation and introduces a click simulation strategy for training. The method in [36] incorporates attention mechanisms into CNN-based architecture for interactive segmentation.

It is worth noting that, human interactions can come in different formats. For instance, early efforts [54] adopt bounding boxes as interaction feedback. Besides, scribbles are also used in the early works [51] to provide more detailed user guidance for precise segmentation. However, drawing a scribble upon the target object poses extra burdens on the user, compared to simpler forms of interaction such as clicks. Thereby, most recent methods adopt clicks as the main form of interaction. In specifics, DEXTR [39] uses extreme points of the target object, i.e. left-most, right-most, top, and bottom pixels as inputs, which is similar to bounding box interaction. Polygon-RNN [7] formulates interactive image segmentation as a polygon prediction problem where a recurrent neural network is used to sequentially predict the vertices of the polygons outlining the object-to-segment. VMN [58] leverages memory-augmented networks to quickly encode past segmentation information and a quality assessment module to estimate the quality of segmentation predictions, enabling efficient retrieval of information and active learning through preferentially labeling the lowest-quality slice for refinement. Clicks on object boundaries are also adopted in [24, 27] as an effective human interaction. There are also some methods [12, 4, 56] that combine different types of interactions to generate robust segmentation masks, e.g., methods in [4, 56] combine bounding boxes and clicks to provide more specific object guidance. Amongst, foreground-background point clicks [45, 37, 9] have gradually become the main interaction way due to the simplicity. We also adopt the foreground-background point clicks as the interaction mode in this work.

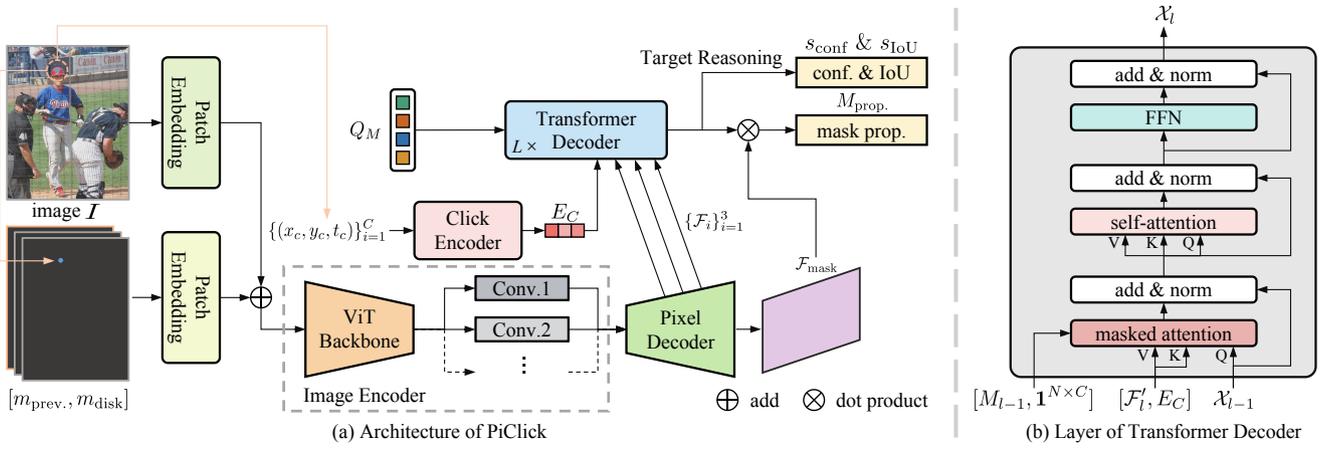

**Figure 3:** The framework of PiClick. (a) First, the input image $\mathcal{I}$, the previous segmentation mask $m_{\text{prev.}}$ and the disk maps of clicks $m_{\text{disk}}$ are fused and fed to the image encoder and pixel decoder to obtain the click-aware multi-scale features $\{\mathcal{F}_i\}_{i=1}^3$ and mask feature $\mathcal{F}_{\text{mask}}$. The user clicks $\{(x_c, y_c, t_c)\}_{i=1}^C$ are encoded into position-aware embeddings $E_C$. Then a set of mask queries $Q_M$ are fed to the Transformer decoder to attend to the click-aware features $\{\mathcal{F}_i\}_{i=1}^3$ with the guidance of the clicks' position-aware embeddings $E_C$. Finally, a mask head is adopted to generate diverse segmentation mask proposals $M_{\text{prop.}}$ for all the mask queries $Q_M$. The Target Reasoning module is designed to predict the confidence scores $s_{\text{conf}}$ of each mask proposal and the IoU $s_{\text{IoU}}$ between the proposals and the ground truth mask, thus automatically picking out the desired one from the mask proposals. (b) Symbol meanings are explained in Sec. 3.1.

To feed clicks to the model, previous click-based interactive methods encode clicks as raw binary representation. Encoding clicks into gaussians or disks with a fixed radius are two commonly used methods of raw binary representation for clicks [45]. However, as shown in Fig. 2(a), when the click is close to the mask boundary, the raw binary representation encoding may unintentionally include both positive and negative regions, hindering the model's ability to learn precise masks. In addition to raw binary representation, PiClick simultaneously encodes clicks as feature-level representations, enabling the model to learn a more detailed mask.

Furthermore, these methods encounter the target ambiguity issue, where one user click may correspond to multiple potential targets in complex scenes instead of a clear-cut single target. As a result, the single-mask methods often generate distorted masks that mix different potential targets, creating confusion and uncertainty in the mask output. For example, as shown in Fig. 2(d), a click on a shoe might also indicate the person wearing it.

### 2.2. Multiple-Output Methods

A few efforts [31, 29] try to explore diversities in the output masks to solve the target ambiguity issue, which we termed as multiple-mask methods. Specifically, these multiple-mask methods employ separate convolutional branches to generate multiple segmentation masks. Limited by the locality of convolutional neural networks, the generated segmentation masks share no inter-mask interactions among each other, thus leading to similar outputs and failing to collaboratively capture different targets. To resolve this lack of diversity, a few methods adopt specific designs to force diversity of the multiple segmentation masks. MultiSeg [31] assigns certain scales to each segmentation branch, forcing each branch to segment objects on a certain scale. Latent Diversity [29] designs diversity loss to encourage difference among outputs, while these outputs are not ensured to be semantically meaningful. Nonetheless, the output masks still tend to be similar, thus post-processing such as non-maximum processing [31] and separate classification networks [29] are required for mask selection, posing heavy computation overhead. In this work, on the contrary, we propose an end-to-end Transformer architecture to automatically generate and select diversified and semantically meaningful masks for click-based interactive segmentation. Specifically, a set of learnable mask queries dynamically encodes the click position-aware embeddings and click-aware image features, and interact with each other in a transformer decoder to ensure diversity, once and for all relieving the needs for diversity loss [29], multi-scale output [31], and non-maximum-suppression [31]. As a result, the training and testing pipeline of interactive segmentation is greatly simplified, achieving more practical value in real annotation applications. Recently, the segmentation foundation model named Segment-Anything (SAM) [26] was proposed. It is trained on large-scale data and generally achieves pretty good performance on various segmentation tasks. However, as it is primarily designed for general purposes rather than being tailored for specific tasks, it may not achieve optimal segmentation results on this specific interactive segmentation task. Our experiments demonstrate that PiClick surpasses SAM in several standard interactive segmentation datasets, particularly in challenging scenarios involving objects with thin and elongated shapes.

## 3. PiClick

The overall framework of our PiClick is shown in Fig. 3. First, PiClick generates multiple diverse segmentation masks with a set of mask queries, introduced in Sec. 3.1. Then, a Target Reasoning module (TRM) predicts the confidence score of each generated proposal and the IoU between generated proposals and the potential ground truth mask to pick out the desired one, described in Sec. 3.2. The training and inference pipelines are illustrated in Sec. 3.3.

### 3.1. Diverse Set Prediction

**Backbone and Pixel Decoder** As shown in Fig. 3, we represent the clicks into a two-channel disk map $m_{\text{disk}} = [m_{\text{disk}}^{pos}, m_{\text{disk}}^{neg}] \in \mathbb{R}^{2 \times H \times W}$ as [37]. The $m_{\text{disk}}^{pos}$ is a binary map where only the pixels within circles centered at positive clicks are set to 1. The $m_{\text{disk}}^{neg}$ is similar to $m_{\text{disk}}^{pos}$ but represents negative clicks. The radius of each circle is set to a certain value. Then we concatenate the disk map and the previous segmentation mask $m_{\text{prev.}} \in \mathbb{R}^{1 \times H \times W}$ (we will discuss in Sec. 3.3), and forward the concatenated map through the Patch Embedding layer to obtain the embedded results. Meanwhile, we forward the image $I \in \mathbb{R}^{3 \times H \times W}$ through another Patch Embedding layer to obtain image embedding.

The image embedding and the embedding of concatenated map are fused and then forwarded through the image encoder to get the feature maps. Note that the image encoder shown in Fig. 3(a) consists of a ViT backbone [13] and a convolutional neck. We use $\mathcal{F}_b \in \mathbb{R}^{HW/16^2 \times D_b}$ to denote the feature maps obtained through the ViT backbone. The output feature $\mathcal{F}_b$ from the ViT backbone is passed through the convolutional neck, which consists of four parallel convolutional layers, to obtain the feature pyramid $\{\mathcal{F}_{n_i}\}_{i=1}^4 = \{\text{Conv}_i(\mathcal{F}_b) \in \mathbb{R}^{H/2^{i+1} \times W/2^{i+1} \times D_i}\}_{i=1}^4$. The structure of pixel decoder is the same as Mask2Former [10]. As a result, the feature pyramid is processed by the pixel decoder to obtain the final click-aware multi-scale features $\{\mathcal{F}_i\}_{i=1}^3$ and mask feature $\mathcal{F}_{\text{mask}}$.

In our framework, the mask proposals are iteratively generated and refined. The training and inference pipelines are detailed in Sec. 3.3.

**Click Encoder.** Given $C$ user clicks $\{(x_c, y_c, t_c)\}_{i=1}^C$, where $(x_c, y_c)$ represents the position of click $c$, and $t_c \in \{0, 1\}$ denotes the type of the clicks (1 for positive click and 0 for negative click). We encode them into position-aware embeddings $E_C \in \mathbb{R}^{C \times D}$:

$$E_c = \text{PE}(x_c, y_c) + t_c E_p + (1 - t_c) E_n. \quad (1)$$

The PE(·) means positional encoding in [46], while $E_p, E_n \in \mathbb{R}^{1 \times D}$ are two learnable embeddings for positive and negative clicks respectively.

The method of encoding clicks as raw binary representation, which requires setting a radius, can potentially restrict the precise localization capability of the model. We enhance the model's localization ability by encoding clicks as feature-level representation, and the effectiveness of this approach is demonstrated in Tab. 3.

**Transformer Decoder.** Then, the learnable mask queries $Q_M \in \mathbb{R}^{N \times D}$ are used to represent multiple potential segmentation masks, which are randomly initialized. Specifically, we feed the mask queries $Q_M$, click position-aware embeddings $E_C$, and the click-aware multi-scale features $\{\mathcal{F}_i\}_{i=1}^3$ to a Transformer decoder, in which click position-aware embeddings $E_C$ guide the mask queries $Q_M$ to attend to the precise target region of click-aware features. The mask queries interact with each other to ensure diverse features and avoid duplicated mask predictions in the self-attention layers. As shown in Fig. 3(b), the cross-attention among mask queries, click position-aware embeddings, and the multi-scale features is defined as

$$\mathcal{X}_l = \text{softmax}(\tilde{\mathcal{M}}_{l-1} + Q_l \mathcal{K}_l^T) \mathcal{V}_l + \mathcal{X}_{l-1}, \quad (2)$$

$$\tilde{\mathcal{M}}_{l-1}(k) = \begin{cases} 0 & \text{if } [M_{l-1}, \mathbf{1}^{N \times C}](k) = 1 \\ -\infty & \text{otherwise} \end{cases}. \quad (3)$$

Here, $l$ is the layer index, $\mathcal{X}_l \in \mathbb{R}^{N \times D}$ refers to $N$ $D$-dimensional query features at the $l^{\text{th}}$ layers and $Q_l = f_Q(\mathcal{X}_{l-1}) \in \mathbb{R}^{N \times D}$, where $\mathcal{X}_0 = Q_M$. $\mathcal{K}_l, \mathcal{V}_l \in \mathbb{R}^{(H_l W_l + C) \times D}$ are the concatenated features $[\mathcal{F}_l', E_C] \in \mathbb{R}^{(H_l W_l + C) \times D}$ ($\mathcal{F}_l' = \mathcal{F}_{l \bmod 3}$) of image features and click position-aware embeddings under transformation $f_K(\cdot)$ and $f_V(\cdot)$ respectively. $M_{l-1} \in \{0, 1\}^{N \times H_l W_l}$ is the binarized output(thresholded at 0.5) of the resized mask prediction of the previous $(l-1)^{\text{th}}$ Transformer decoder layer, and the concatenated attention mask $[M_{l-1}, \mathbf{1}^{N \times C}] \in \mathbb{R}^{N \times (H_l \times W_l + C)}$. $M_0$ is the binary mask prediction obtained from $\mathcal{X}_0$.

After the Transformer decoder, the mask queries dynamically encode diverse click-guided features for different semantical meaningful masks. A unified segmentation head $\mathcal{H}_s$ containing three MLP layers is applied to generate the set of segmentation masks $M_{\text{prop.}} = \{m_i\}_{i=1}^N \in \mathbb{R}^{N \times H \times W}$, which we name as mask proposals, and each mask $m_i$ is obtained by

$$m_i = \mathcal{F}_{\text{mask}} \otimes \mathcal{H}_s(\mathcal{X}_i^{-1}), \quad (4)$$

where the $\mathcal{H}_s(\mathcal{X}_i^{-1})$ indicates the dot product weight generated by mask query of the last Transformer decoder layer $\mathcal{X}_i^{-1}$, and the $\otimes$ means dot product operation. With the segmentation head, each mask query generates a segmentation mask proposal based on the user clicks.

### 3.2. Target Reasoning Module

After the multiple mask proposals are generated, the user is supposed to pick out the desired mask, which also requires user interactions. Therefore, we propose an effective Target Reasoning module (TRM) to automatically pick out the target mask for more efficient user interaction. TRM predicts the confidence scores of each mask proposal and the IoU between each mask proposal and ground truth, which are calculated as

$$s_{\text{conf},i} = \mathcal{H}_{\text{conf}}(\mathcal{X}_i^{-1}), s_{\text{IoU},i} = \mathcal{H}_{\text{IoU}}(\mathcal{X}_i^{-1}). \quad (5)$$

**Algorithm 1:** One iteration for training PiClick

**Input:** Image $\mathcal{I} \in \mathbb{R}^{H \times W \times 3}$, all annotated masks $\hat{M} = \{\hat{m}_i\}_{i=1}^{N_M}$, learnable parameters $\theta$ for PiClick, previous predicted mask $m_{\text{prev.}} = \mathbf{0}^{1 \times H \times W}$, and a sampled primary target mask $\hat{m}_p \in \hat{M}$.

```
// (1) Click-and-Mask Simulation: random click strategies
```
1   $n_{\text{init}} = \text{random.randint}(1, 10)$ `// randomly chosen integer n_init in the range [1,10]`
2   $C = \{(x_i, y_i, t_i)\}_{i=1}^{n_{\text{init}}} = \text{generate}(\hat{m}_p, \hat{M})$ `// randomly generate clicks C according to primary target mask m̂_p ∈ M̂`
```
// (1) Click-and-Mask Simulation: interactive click simulation strategies
```
3   $n_{\text{inter}} = \text{random.randint}(0, 4)$ `// randomly chosen integer n_inter in the range [0,4].`
4   **for** $i = 1; i \leq n_{\text{inter}}; i = i + 1$ **do**
5      $M_{\text{prop.}} = \{m_i\}_{i=1}^{N} = \text{PiClick}_\theta(\mathcal{I}, C, m_{\text{prev.}})$ `// predict N mask proposals M_prop. with PiClick`
6      $m_{\text{prev.}} = m_{\text{chosen}} = \text{random.choice}(M_{\text{prop.}})$ `// randomly select one mask from M_prop.`
7      $C = C + \{(x_a, y_a, t_a)\}$ `// place next click (x_a, y_a, t_a) on the erroneous region of m_chosen`
```
// (2) Forward: predict N mask proposals M_prop., confidence scores s_conf, and IoU predictions s_IoU with PiClick
```
8   $M_{\text{prop.}}, s_{\text{conf}}, s_{\text{IoU}} = \{m_i\}_{i=1}^{N}, \{s_{\text{conf},i}\}_{i=1}^{N}, \{s_{\text{IoU},i}\}_{i=1}^{N} = \text{PiClick}_\theta(\mathcal{I}, C, m_{\text{prev.}})$
```
// (3) Backward: generate multiple feasible target masks M̂_C ⊆ M̂; generate pseudo IoU labels ŝ_IoU for
   predicted segmentation masks M_prop. and confidence score ŝ_conf according to Hungarian algorithm
```
9   $\hat{s}_{\text{IoU}} = \{\hat{s}_{\text{IoU},i}\}_{i=1}^{N} = \text{IoU}(M_{\text{prop.}}, \hat{m}_p)$
10   $\mathcal{L}_{\text{match}} = \mathcal{L}_{\text{dice}}(M_{\text{prop.}}, \hat{M}_C) + \mathcal{L}_{\text{focal}}(M_{\text{prop.}}, \hat{M}_C) + \mathcal{L}_{\text{L1}}(s_{\text{IoU}}, \hat{s}_{\text{IoU}}) + \mathcal{L}_{\text{BCE}}(s_{\text{IoU}}, \hat{s}_{\text{IoU}})$ `// calculate loss`
11   $\theta \leftarrow \theta - \eta \nabla_\theta \mathcal{L}_{\text{match}}$ `// update Piclick`

**Output:** optimized model $\text{PiClick}_\theta$.

Here, the confidence score head $\mathcal{H}_{\text{conf}}$ and the IoU prediction head $\mathcal{H}_{\text{IoU}}$ containing three MLP layers respectively.

TRM is based on the assumption that the objects that users focus on are partially predictable based on the relative locations of clicks. For instance, when a click is placed on the hat, it could potentially indicate either the hat itself or the person wearing the hat. However, in practice, users often tend to initially click the central region of the target and then proceed to click unexpected segments in subsequent interactions. As a result, the click on the hat is more likely to signify the hat rather than the person.

MultiSeg [31] ranks the predicted masks by the corresponding mask confidence scores. LatentDiversity [29] additionally optimizes a binary classification network to predict the confidence score of selecting each predicted mask. However, compared with predicting the potential IoU, it is much harder to directly predict how well each mask matches the human intention (*i.e.,* the mask confidence), rendering previous mask-confidence-based selection less effective.

Therefore, we propose to compute the IoU between each output mask and the target ground truth as the objective of TRM. Specifically, with the predicted mask set, we first compute the intersection over union between each predicted mask and the target ground truth mask as the supervision. Then we utilize Equation 5 to reason about the IoU between each mask query and the ground truth. In practice, we find that appropriately combining both of them yields the best selection results. We think it is because the mask confidence score may be still useful for excluding apparent false cases. We experimentally show our IoU-based TRM obtains similar results to human selection, which illustrates the strong capability of our method to mimic human interaction during mask selection.

### 3.3. Training and Inference

**Training.** We show one iteration for training PiClick in Algorithm 1. In each training iteration, the process includes click-and-mask simulation (lines 1-7 in Algorithm 1), forward pass (line 8 in Algorithm 1), and backward pass (lines 9-10 in Algorithm 1). Given a training sample $(\mathcal{I}, \hat{M})$, where the $\mathcal{I}$ denotes the input image and the $\hat{M} = \{\hat{m}_i\}_{i=1}^{N_M}$ denotes all the ground-truth segmentation masks of the image. We first sample a primary target mask $\hat{m}_p$ out of $\hat{M}$, which represents the user intentions. During the click-and-mask simulation, we automatically simulate clicks based on the current segmentation and gold standard segmentation. Specifically, we use a combination of random click simulation (lines 1-2 in Algorithm 1) and interactive click simulation strategies (lines 3-7 in Algorithm 1), following the pipelines in SimpleClick [37]. For the random click strategy, we generate $n_{\text{init}}(1 \leq n_{\text{init}} \leq 10)$ clicks randomly to match the positions of the primary target mask $\hat{m}_p$, without considering the order of the clicks. For the interactive click simulation strategies, we simulate $n_{\text{inter}}$ interactions of clicks. For this round, we randomly select one mask from $N$ mask proposals as the prediction result (empty mask in the first interaction). Here, $n_{\text{inter}}$ is a randomly chosen integer ranging from 0 to 4. Each current click is strategically placed in the area where the previous interaction's prediction mask $m_{\text{prev.}}$ is incorrect. Once we have established the $n_{\text{init}} + n_{\text{inter}}$ clicks and the previous prediction mask $m_{\text{prev.}}$, we select the feasible target masks $\hat{M}_C \subseteq \hat{M}$ for supervision which satisfy the constraint that the positive clicks are inside $\hat{M}_C$ and the negative clicks are outside $\hat{M}_C$.

After click-and-mask simulation, we feed the image $\mathcal{I}$, the user clicks $\mathcal{C}$, and the previous segmentation mask $m_{\text{prev.}}$ from the last interaction, into the PiClick Network. The PiClick generates a set of mask proposals $M_{\text{prop.}} = \{m_i\}_{i=1}^N$, corresponding IoU predictions $s_{\text{IoU}} = \{s_{\text{IoU},i}\}_{i=1}^N$, and corresponding confidence scores $s_{\text{conf}} = \{s_{\text{conf},i}\}_{i=1}^N$, where $N$ is the number of the mask queries in PiClick. The pseudo IoU labels $\hat{s}_{\text{IoU}}$ are generated by computing the IoU between each predicted mask and the primary target ground truth, and the confidence scores $\hat{s}_{\text{conf}} = \{0, 1\}^N$ are derived from the results of matching with the multiple feasible target masks $\hat{M}_C$ using the Hungarian algorithm (predicted matches are assigned a value of 1; non-matches are assigned a value of 0.). Then we adopt similar training loss in DETR [6] for the set prediction:

$$\mathcal{L}_{\text{match}} = \begin{array}{l} \mathcal{L}_{\text{dice}}(M_{\text{prop.}}, \hat{M}_C) + \mathcal{L}_{\text{focal}}(M_{\text{prop.}}, \hat{M}_C) \\ + \mathcal{L}_{L1}(s_{\text{IoU}}, \hat{s}_{\text{IoU}}) + \mathcal{L}_{\text{BCE}}(s_{\text{conf}}, \hat{s}_{\text{conf}}) \end{array}. \quad (6)$$

We adopt Dice loss [41] and binary focal loss loss [33] as mask loss, adopt the L1 loss as IoU loss, and adopt binary cross entropy loss as confidence loss.

**Inference.** During the inference, the user first adds a single click $\mathcal{C}$ upon the target object. PiClick predicts the multiple potential masks, and corresponding IoUs and confidence scores for each mask. The $j^{th}$ prediction mask is automatically selected, where $j = \arg\max_i(s_{\text{IoU},i} \times s_{\text{conf},i})$. After that, if the picked mask is close enough, then we obtain the final segmentation result. If the picked mask is not close enough, the user could adjust the mask by (i) replacing it with another proposal mask possessing higher IoU with the target ground truth or (ii) adding more clicks based on the picked mask to conduct the refinement for one more interaction.

## 4. Experiments

### 4.1. Experimental Setups

**Training Datasets.** Following RITM [45] and SimpleClick [37], we conduct experiments on 10 public datasets including 7 natural datasets and 3 medical datasets. We use COCO [34]+LVIS [20] (C+L) for training (a combination of COCO [34] and LVIS [20] dataset), where COCO contains 118K training images (1.2M instances), and LVIS shares the same images with COCO but has more instance masks and higher mask quality.

**Test Datasets.** Following [37, 45, 14], we evaluate our model on 9 benchmarks: GrabCut [43], Berkeley [40], DAVIS [42], Pascal VOC [15], COCO MVal [34], SBD [21], ssTEM [17], BraTS [3], and OAIZIB [2].

**Evaluation Metrics** Following previous works [37, 45], we perform the evaluation in terms of the standard Number of Clicks (NoC), measuring the number of clicks required to achieve the predefined Intersection-over-Union (IoU) between the predicted and ground-truth masks. The NoC with IoU threshold 85% and 90% are denoted by NoC%85 and NoC%90 respectively. The maximum number of clicks for each instance is set to 20. Following SimpleClick [37], we also evaluate the average IoU given $k$ clicks (mIoU@$k$) to measure the segmentation quality given a fixed number of clicks.

**Implementation Details** In the experiments, we adopt vanilla ViT model [13] as backbones, which is initialized by MAE [22] pre-training on ImageNet [11]. The convolutional neck, Transformer decoder, Target Reasoning module (TRM), and mask head are randomly initialized. For ViT-B with the convolutional neck, $\{D_i\}_{i=1}^4 = \{256, 256, 256, 256\}$, while for ViT-L model with the neck, $\{D_i\}_{i=1}^4 = \{192, 384, 768, 1536\}$. We train our model on the COCO+LVIS dataset for 62 epochs with Adam optimizer. The parameters are set to $\beta_1 = 0.9$, $\beta_2 = 0.999$. The initial learning rate is set to $5.15 \times 10^{-5}$ (ViT-B) and $6.42 \times 10^{-5}$ (ViT-L), and decays by a multiplicative factor of 0.1 at the 50th epoch. The batch size is set to 136 (ViT-B) and 56 (ViT-L). Our model is implemented in PyTorch and trained with 8 NVIDIA Tesla V100 GPUs (32 GB memory). We use the following data augmentation techniques during training: random resizing (scale range from 0.75 to 1.25), random flipping and rotation, random brightness contrast, and random cropping. Following SimpleClick [37], we finetune our model with 448×448 images and non-shifting window attention for better performance.

### 4.2. Comparison with state-of-the-art methods

We compare PiClick with previous state-of-the-art methods in Tab. 1. As shown, PiClick achieves the best performance on all the natural benchmarks. We also compare PiClick with previous methods in terms of mIoU@1 on 6 natural datasets in Fig. 4(a). PiClick (ViT-B) outperforms all previous methods by above 3.55% mIoU@1. As PiClick and the SOTA method SimpleClick [37] share the same backbone and training datasets, we conduct detailed comparisons with SimpleClick to demonstrate the superiority of our proposed PiClick. To be specific, PiClick (ViT-B) significantly outperforms SimpleClick (ViT-B) by 5.44% considering mIoU@1 on 6 natural datasets. This result suggests that PiClick can effectively mitigate the target ambiguity issue, which often arises when the user clicks are insufficient to precisely define the target objects. The visualizations comparison between PiClick and SimpleClick are shown in Fig. 5, in which the sub-figures demonstrate the cases with target ambiguity issues. For example, as shown in the second row of Fig. 5, the green click may refer to the backpack, the person, or a combination of the person and backpack. The performance of SimpleClick is restricted by the target ambiguity issue which may result in unexpected mask, while PiClick can mitigate such issue by first generating all potential masks and picking out the desired one with TRM. Fig. 4(c) shows the training curve on 6 natural datasets. We observe that our PiClick converges faster than SimpleClick with the same backbone and the other typical click-based methods. To sum up, our PiClick (ViT-B) achieves the best mIoU@1 performance on all benchmarks compared with previous state-of-the-arts, which verifies the effectiveness of our proposed method.

**Table 1**
**NoC performance comparison with existing methods.** We report results on 6 nature benchmarks: GrabCut [43], Berkeley [40], SBD [21], DAVIS [42], Pascal VOC [15], and COCO MVal [34]. The best results are set in bold. † denotes the generalist model. ‡ denotes a number provided by SEEM [59]. ℘ denotes the results of training for 62 epochs, used for the fair comparison.

| Method | Backbone | GrabCut | | Berkeley | | SBD | | DAVIS | | Pascal VOC | | COCO MVal | |
|---|---|---|---|---|---|---|---|---|---|---|---|---|---|
| | | NoC85 | NoC90 | NoC85 | NoC90 | NoC85 | NoC90 | NoC85 | NoC90 | NoC85 | NoC90 | NoC85 | NoC90 |
| RITM [45] | HRNet-18 | 1.42 | 1.54 | - | 2.26 | 3.80 | 6.06 | 4.36 | 5.74 | 2.28 | - | - | 2.98 |
| RITM [45] | HRNet-32 | 1.46 | 1.56 | 1.43 | 2.10 | 3.59 | 5.71 | 4.11 | 5.34 | 2.19 | 2.57 | - | 2.97 |
| CDNet [8] | ResNet-34 | 1.40 | 1.52 | 1.47 | 2.06 | 4.30 | 7.04 | 4.27 | 5.56 | 2.74 | 3.30 | 2.51 | 3.88 |
| PseudoClick [38] | HRNet-32 | 1.36 | 1.50 | 1.40 | 2.08 | 3.46 | 5.54 | 3.79 | 5.11 | 1.94 | 2.25 | - | - |
| FocalClick [9] | SegF-B0 | 1.40 | 1.66 | 1.59 | 2.27 | 4.56 | 6.86 | 4.04 | 5.49 | 2.97 | 3.52 | 2.65 | 3.59 |
| FocalClick [9] | SegF-B3 | 1.44 | 1.50 | 1.55 | 1.92 | 3.53 | 5.59 | 3.61 | 4.90 | 2.46 | 2.88 | 2.32 | 3.12 |
| SAM† [26] | ViT-B | - | - | - | - | 6.50‡ | 9.76‡ | - | - | 3.30‡ | 4.20‡ | - | - |
| SEEM† [59] | DaViT-B | - | - | - | - | 6.67 | 9.99 | - | - | 3.41 | 4.33 | - | - |
| InterFormer [23] | ViT-B | 1.38 | 1.50 | 1.99 | 3.14 | 3.78 | 6.34 | 4.10 | 6.19 | - | - | - | - |
| SimpleClick [37] | ViT-B | 1.38 | 1.48 | **1.36** | 1.97 | 3.43 | 5.62 | 3.66 | 5.06 | 2.06 | 2.38 | 2.16 | 2.92 |
| SimpleClick℘ [37] | ViT-B | 1.54 | 1.64 | 1.40 | 1.99 | 3.42 | 5.58 | 3.60 | 5.09 | 2.09 | 2.43 | 2.13 | 2.88 |
| AdaptiveClick [32] | ViT-B | 1.34 | 1.48 | 1.40 | 1.83 | 3.29 | 5.40 | 3.39 | 4.82 | 2.03 | 2.31 | - | - |
| VPUFormer [57] | ViT-B | 1.34 | 1.40 | 1.38 | **1.71** | 3.32 | 5.45 | 3.48 | 4.82 | - | - | - | - |
| PiClick | ViT-B | **1.26** | **1.37** | **1.36** | 1.78 | **3.11** | **5.32** | 3.44 | **4.60** | **1.86** | **2.23** | **2.01** | **2.76** |
| InterFormer [23] | ViT-L | 1.26 | 1.36 | 1.61 | 2.53 | 3.25 | 5.51 | 4.54 | 5.21 | - | - | - | - |
| SimpleClick [37] | ViT-L | 1.32 | 1.40 | 1.34 | 1.89 | 2.95 | 4.89 | 3.26 | 4.81 | 1.72 | **1.96** | 2.01 | 2.66 |
| SimpleClick℘ [37] | ViT-L | 1.46 | 1.52 | 1.41 | 1.93 | 2.94 | 4.87 | 3.30 | 4.78 | 1.79 | 2.04 | 2.00 | 2.69 |
| PiClick | ViT-L | **1.22** | **1.30** | **1.29** | **1.65** | **2.86** | **5.00** | **3.18** | **4.52** | **1.71** | 1.97 | **1.91** | **2.57** |

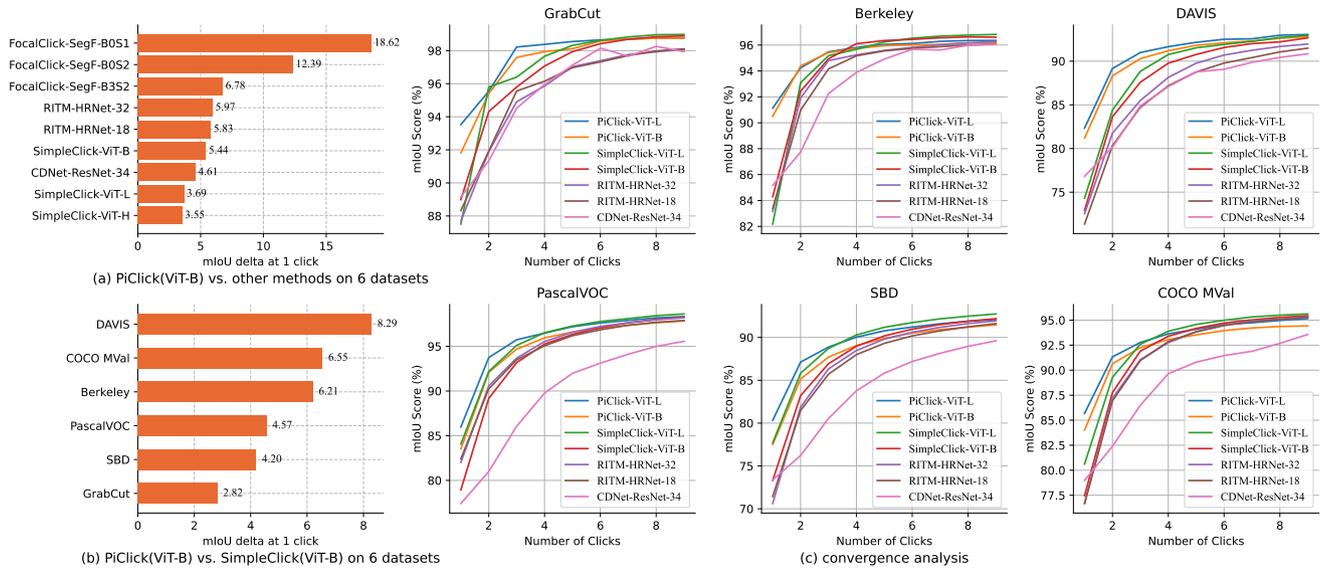

**Figure 4: mIoU performance comparison with existing methods**, We report results on 6 nature benchmarks: GrabCut [43], Berkeley [40], SBD [21], DAVIS [42], Pascal VOC [15], and COCO MVal [34]. (a) mIoU delta between our PiClick and the previous state-of-the-art methods in 6 datasets; (b) mIoU delta between PiClick and the SimpleClick on all 6 datasets; (c) mIoU varying with the number of click points on 6 natural datasets.

Concurrently, some general-purpose foundation models, such as SAM (segment-anything) [26] and SEEM [59], came out and gained a lot of attention. Therefore we also compare PiClick with SAM and SEEM in Tab. 1. We observe that the results of SAM and SEEM performs even worse than the existing methods, although they are trained with a super large amount of images. While SAM is designed for interactive segmentation, it focuses on segmenting objects at any granularity, which may result in lower segmentation quality for smaller objects, as illustrated in Fig. 9.

### 4.3. Out-of-Domain Evaluation

To test the generalization ability of our proposed method, we further evaluate the PiClick on three medical image datasets: ssTEM, BraTS, and OAIZIB. Tab. 2 and Fig. 6 report the result comparisons on those three datasets. Specifically, Fig. 6(a) compares the per-model mIoU@1 results on three medical datasets, while Fig. 6(b-d) shows the convergence curve on ssTEM, BraTS, and OAIZIB. Overall, PiClick demonstrates state-of-the-art performance compared to existing methods, showcasing its strong generalization ability to handle out-of-domain images, *e.g.*, medical images. Note that PiClick (ViT-B) achieves 62% mIoU@1 on the BraTS dataset, significantly outperforming the previous best score by 47%. The higher target ambiguity in the BraTS dataset, where many target masks have overlapping regions, makes it challenging to determine the specific target with a single click, as shown in Fig. 7. The BraTS dataset is collected specifically for the segmentation of brain tumors. Methods that utilize a single-mask approach tend to produce masks covering the entire brain as the outcome of the initial round of interactive segmentation, which leads to a relatively low mIoU@1 at only 15%. In contrast, our method generates multiple potential outcomes and tends to avoid selecting the mask that encompasses the entire area, which enables PiClick to significantly enhance performance on the BraTS dataset. In contrast, ssTEM and OAIZIB datasets have fewer ambiguity cases, which accounts for PiClick achieving comparable performance to existing methods in these datasets.

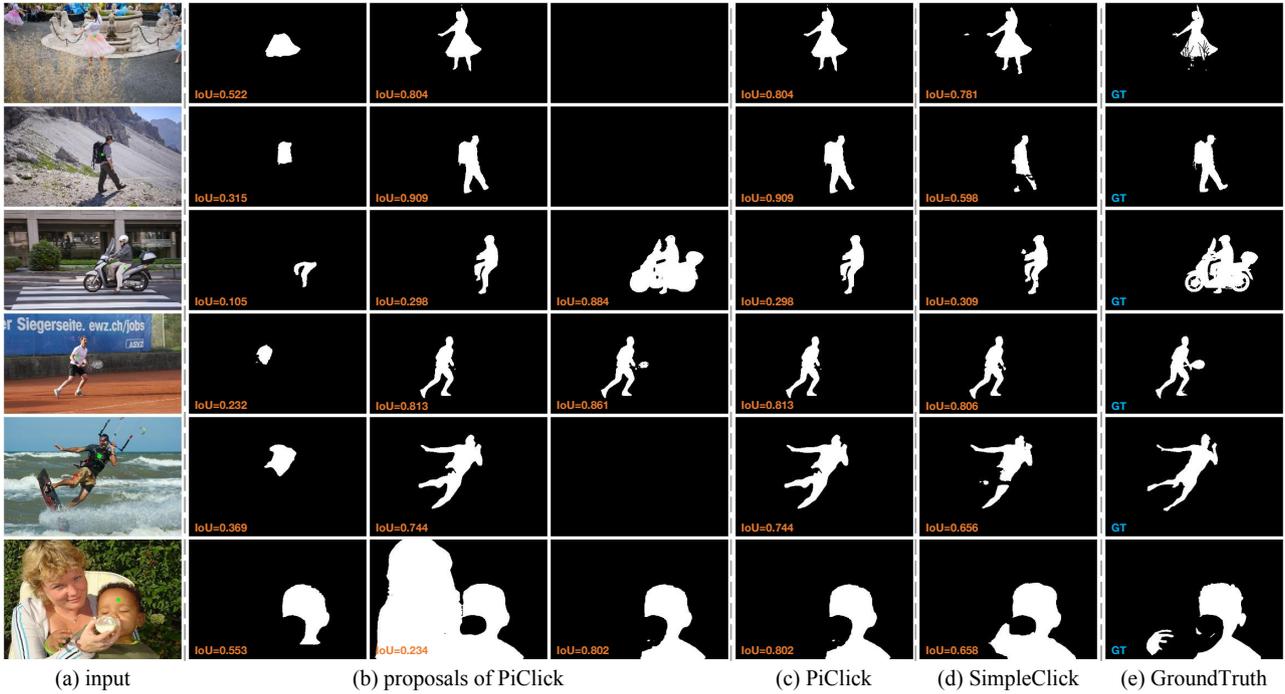

(a) input     (b) proposals of PiClick     (c) PiClick     (d) SimpleClick     (e) GroundTruth

**Figure 5: Visualization.** (a) Images with one positive click. (b) The segmentation proposals generated by PiClick (ViT-B). (c) Segmentation mask generated by PiClick (ViT-B). (d) Segmentation mask generated by SimpleClick (ViT-B). (e) Ground truth. IoU represents the Intersection over Union between the predicted mask and the ground truth.

### 4.4. Ablation Study

We conduct extensive experiments on 6 natural datasets to demonstrate the effectiveness of each module of PiClick.

**Table 2**
mIoU@1 and NoC%80 comparison on three medical image datasets.

| Method | Backbone | mIoU@1 | | | NoC80 | | |
|---|---|---|---|---|---|---|---|
| | | ssTEM | BraTS | OAIZIB | ssTEM | BraTS | OAIZIB |
| RITM | HRNet-18 | 34.91 | 8.55 | 13.48 | 2.50 | 5.91 | 8.20 |
| RITM | HRNet-32 | 35.25 | 8.72 | 19.51 | 2.45 | 5.63 | 7.77 |
| CDNet | ResNet-34 | 27.75 | 15.21 | 16.32 | 3.60 | 7.05 | 9.31 |
| FocalClick | SegF-B0S1 | 29.39 | 10.04 | 30.52 | 3.75 | 7.45 | 9.04 |
| FocalClick | SegF-B0S2 | 25.35 | 11.14 | **30.93** | 3.85 | 6.55 | 8.27 |
| FocalClick | SegF-B3S2 | 28.07 | 13.18 | 28.04 | 2.75 | 5.28 | **6.87** |
| SimpleClick | ViT-B | 28.48 | 9.70 | 17.28 | 3.40 | 6.24 | 8.43 |
| SimpleClick | ViT-L | 7.66 | 12.06 | 11.48 | 3.58 | 5.35 | 7.97 |
| PiClick | ViT-B | 39.59 | 52.28 | 25.21 | 2.21 | 4.62 | 10.43 |
| PiClick | ViT-L | **58.74** | **62.25** | 21.68 | **1.97** | **4.41** | 9.99 |

**Table 3**
Ablation study on Click Encoder in terms of NoC%90. VOC denotes Pascal VOC and MVal denotes COCO MVal.

| raw | feature- | GrabCut | Berkeley | SBD | DAVIS | VOC | MVal |
|---|---|---|---|---|---|---|---|
| ✓ | | 1.42 | 1.83 | 5.57 | 5.01 | 2.31 | 2.91 |
| | ✓ | 2.04 | 2.93 | 9.51 | 6.64 | 4.18 | 4.83 |
| ✓ | ✓ | 1.37 | 1.78 | 5.32 | 4.60 | 2.23 | 2.76 |

**Multiple Mask vs. Single Mask** Tab. 4 shows the performance of PiClick with different numbers of mask queries on 6 natural datasets. As shown, the performance of PiClick with multiple mask queries outperforms the PiClick with only one mask query by a large margin. The performance improves with an increasing number of mask queries, converging when set to 7. Hence, we adopt 7 mask queries for PiClick. Since the number of masks is related to the data distribution of the actual test dataset, we recommend the practitioners choosing different numbers of mask queries for different cases.

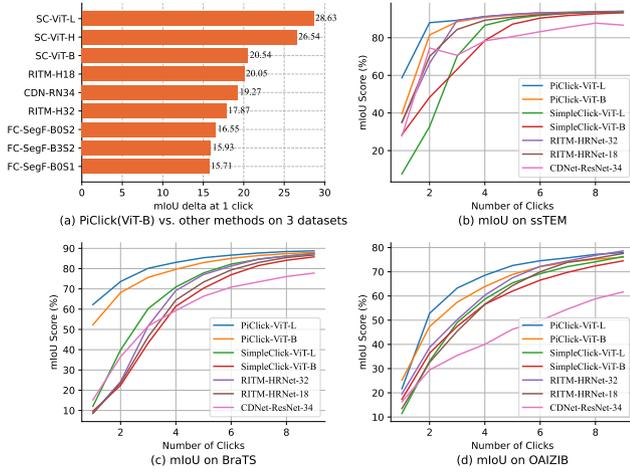

**Table 4**
**Ablation study on the number of mask queries in terms of NoC%90**.

| Number | GrabCut | Berkeley | SBD | DAVIS | VOC | MVal |
|---|---|---|---|---|---|---|
| 1 | 1.64 | 2.23 | 6.55 | 5.72 | 2.85 | 3.60 |
| 3 | 1.36 | **1.77** | 5.74 | 5.09 | 2.35 | 3.03 |
| 5 | **1.35** | 1.87 | 5.65 | 4.90 | 2.26 | 2.82 |
| 7 | 1.37 | 1.78 | **5.32** | **4.60** | **2.23** | **2.76** |
| 9 | 1.41 | 1.83 | 5.40 | 4.78 | 2.27 | 2.85 |

**Table 5**
**Ablation study on the Target Reasoning Module**. The numbers in the table show the additional human selections made for correcting the target mask.

| Conf. | IoU | GrabCut | Berkeley | SBD | DAVIS | VOC | MVal |
|---|---|---|---|---|---|---|---|
| ✓ |   | 0.24 | 0.27 | 0.57 | 0.61 | 0.56 | 0.62 |
|   | ✓ | **0.11** | 0.15 | 0.47 | 0.44 | 0.46 | 0.51 |
| ✓ | ✓ | **0.11** | **0.13** | **0.44** | **0.39** | **0.40** | **0.47** |

**Figure 6: mIoU comparison on 3 medical datasets**: ssTEM, BraTS and OAIZIB. (a) Mean IoU delta between PiClick and the previous state-of-the-art methods on 3 medical datasets; (b-d) mIoU varying with the number of click points on 3 medical datasets.

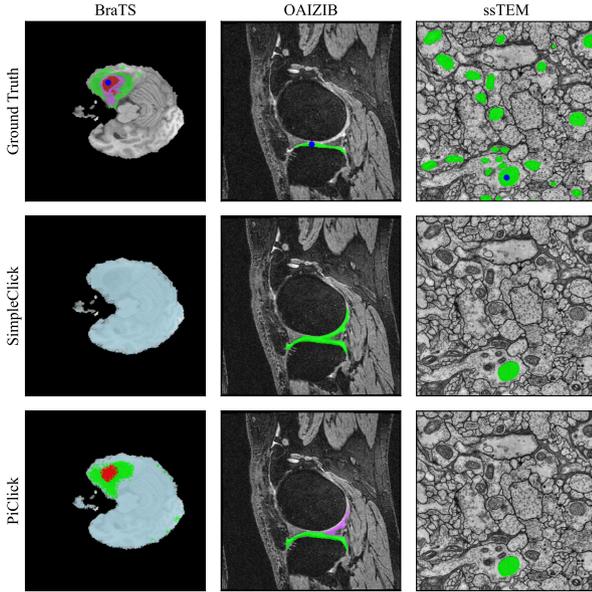

**Figure 7: Visualization on 3 medical datasets**: ssTEM, BraTS and OAIZIB. The BraTS dataset possesses more target ambiguities.

**Effect of Click Encoding** PiClick encodes clicks with both raw binary representation (*i.e.*, the disk map) and feature-level representation (*i.e.*, the output of the click encoder). In Tab. 3, we report the comparisons between the raw binary representation and feature-level representation. Tab. 3 clearly demonstrates that combining both encoding methods significantly outperforms using a single encoding method by a substantial margin. The reason is that when the click is located near the boundary, the raw binary representation, which requires setting a radius, may inadvertently encompass both the positive and negative regions, thus impeding the model's ability to learn fine-grained masks, particularly in scenarios involving small masks. This restriction can potentially limit the model's precise localization capability. Encoding clicks into the position-aware embeddings alleviates the confusion caused by the raw binary encoding method, yielding a significant improvement of model's localization accuracy.

**Effect of Target Reasoning Module** In this work, we design TRM to predict the Intersection over Union (IoU) between each mask proposal and the potential ground truth to identify the most suitable one, aiming to align with human intentions. The combination of the predicted IoU and the mask confidence score is employed to select the final mask. In this ablation study, we compare three different ways to select the potential mask, as shown in Tab. 5. The "Conf." means we directly use the mask confidence score for ranking and selection, following MultiSeg [31] and LatentDiversity [29], The "IoU" means the ranking and selection are based on the IoU prediction of each mask. We also present the results by multiplying the two scores. As shown, TRM with IoU prediction outperforms that with confidence score on all the natural datasets. This implies the mask confidence score may not reflect the human selection preference or intention. We observe the way combining mask confidence score with IoU prediction outperforms that using IoU prediction only. Thus, in our framework, we select the desired mask by multiplying the mask confidence score and the IoU prediction.

**Visualization of predicted IoU and confidence score** Fig. 8 shows the predicted masks and corresponding predicted IoU for different click locations. Purely mask-confidence-based selection doesn't work very well as mask confidence score merely reflects the strength of correlation between predicted mask and the ground-truth, without the ability to infer the user's preference. For example, in Fig. 8(a,2), a click on the hat yields similar mask confidence score for the hat mask and the mask of entire person as the former is part of the latter and they have quite strong correlations. TRM predicts the potential IoU between the predicted mask and the GT mask. (i) From Fig. 8(a,2), it can be observed that although a positive click on the hat corresponds to masks for both the hat and the person, the hat has a higher predicted IoU. This is likely because the positive click position on the hat is closer to the center of the hat and further from the center of the person, making the hat a more likely choice relative to the person. The similar phenomenon can be observed in Fig. 8(b,3) and Fig. 8(d,3). (ii) From Fig. 8(d,2), it is noted that a positive click located at the junction between two objects makes the model more inclined to consider the masks of both objects as likely selections. The same observation can be seen in Fig. 8(b,3) and Fig. 8(c,3). (iii) As shown in Fig. 8(c,1) and Fig. 8(d,1), when a positive click is associated with multiple potential masks and the distance between this click and the centroids of these masks is relatively small, the model often does not choose the largest mask.

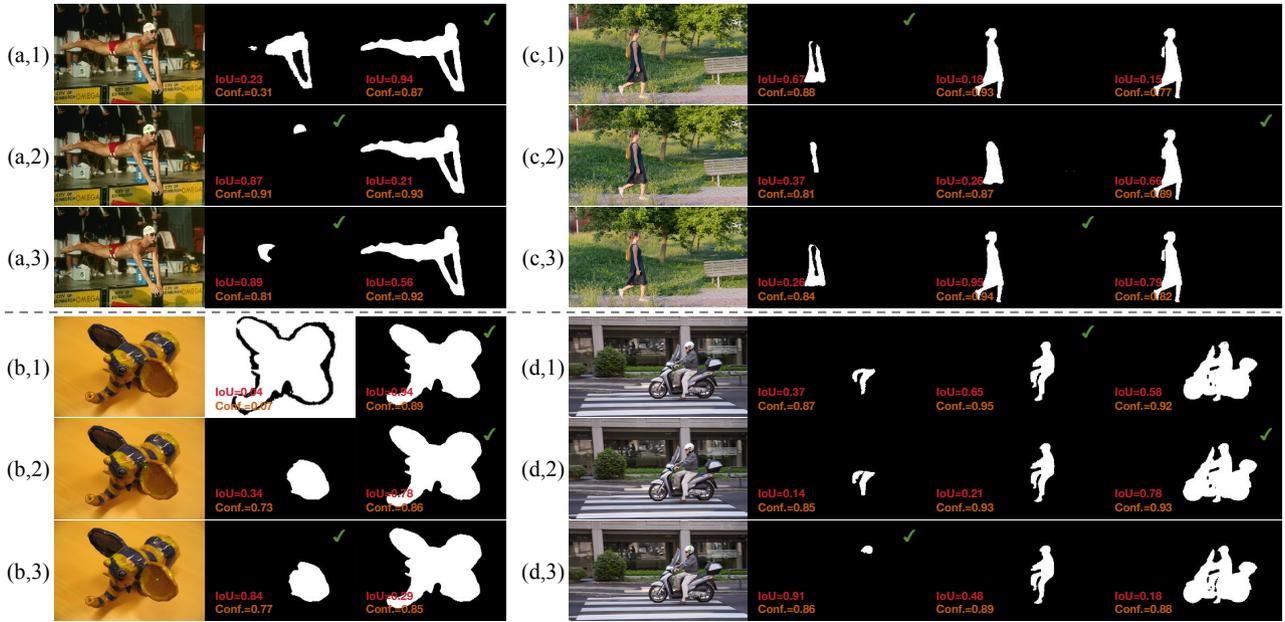

**Figure 8:** Visualization of predicted IoU and mask confidence score. IoU denotes predicted Intersection over Union. Conf. denotes predicted mask confidence score. The check mark denotes the mask selected by our method.

**Efficiency of Our Method** As shown in Tab. 6, the interactive segmentation performance of the multi-mask mask decoder surpasses that of the single mask approach. Moreover, this enhancement does not significantly increase the model's parameter count or inference costs.

**Effect of randomly selecting $m_{\text{prev}}$.** In Algorithm 1 (line 6), we randomly pick a mask proposal as the previous mask $m_{\text{prev}}$. An alternative way is that we select the mask proposal which has the largest IoU with the ground truth as $m_{\text{prev}}$ for training. In this way, for training, we may select the intermediate mask that matches true user intention. But for inference, we have to perform the selection based on model predictions. The distribution shift, which means the model prediction may not always align with the ground truth (or user intention), may constrain the segmentation performance. In contrast, a random selection may mimic diverse user behaviors for the model to learn. As shown in Tab. 7, our way ("Random") performs better than the way ("Largest-IoU") that selects intermediate mask via ground truth.

**Table 6**
Efficiency comparison. † denotes the average NoC across the 6 natural image datasets. ‡ denotes the time cost per click for prediction.

| Method | Backbone | Params. | FLOPs | NoC85† | NoC90† | Cost‡ |
|---|---|---|---|---|---|---|
| SimpleClick | ViT-B | 98M | 162G | 2.34 | 3.24 | 0.02s |
| SimpleClick | ViT-L | 324M | 526G | 2.10 | 2.94 | 0.03s |
| PiClick | ViT-B | 116M | 215G | 2.17 | 3.01 | 0.05s |
| PiClick | ViT-L | 383M | 740G | 2.03 | 2.83 | 0.07s |

**Table 7**
NoC%90 performance comparison. The "Largest-IoU" denotes that during training, we select the mask proposal which has the highest IoU with ground truth as the previous mask. The "Random" denotes that a mask proposal is randomly picked as the previous mask.

| | GrabCut | Berkeley | SBD | DAVIS | VOC | MVal |
|---|---|---|---|---|---|---|
| Largest-IoU | 1.41 | 1.84 | 5.41 | 5.01 | 2.29 | 2.83 |
| Random | 1.37 | 1.78 | 5.32 | 4.60 | 2.23 | 2.76 |

### 4.5. Discussions
Our PiClick may struggle to predict high-quality masks in challenging scenarios, such as objects with thin and elongated shapes or cluttered occlusions (Fig. 9 (a)). Existing methods tried to solve this issue with global and local views [35], while no works by now can solve the quality issue and the target ambiguity issue at the same time. Thus, the interactive segmentation results for the scenario with high-quality mask requirements are still not satisfactory. We remain this as future work.

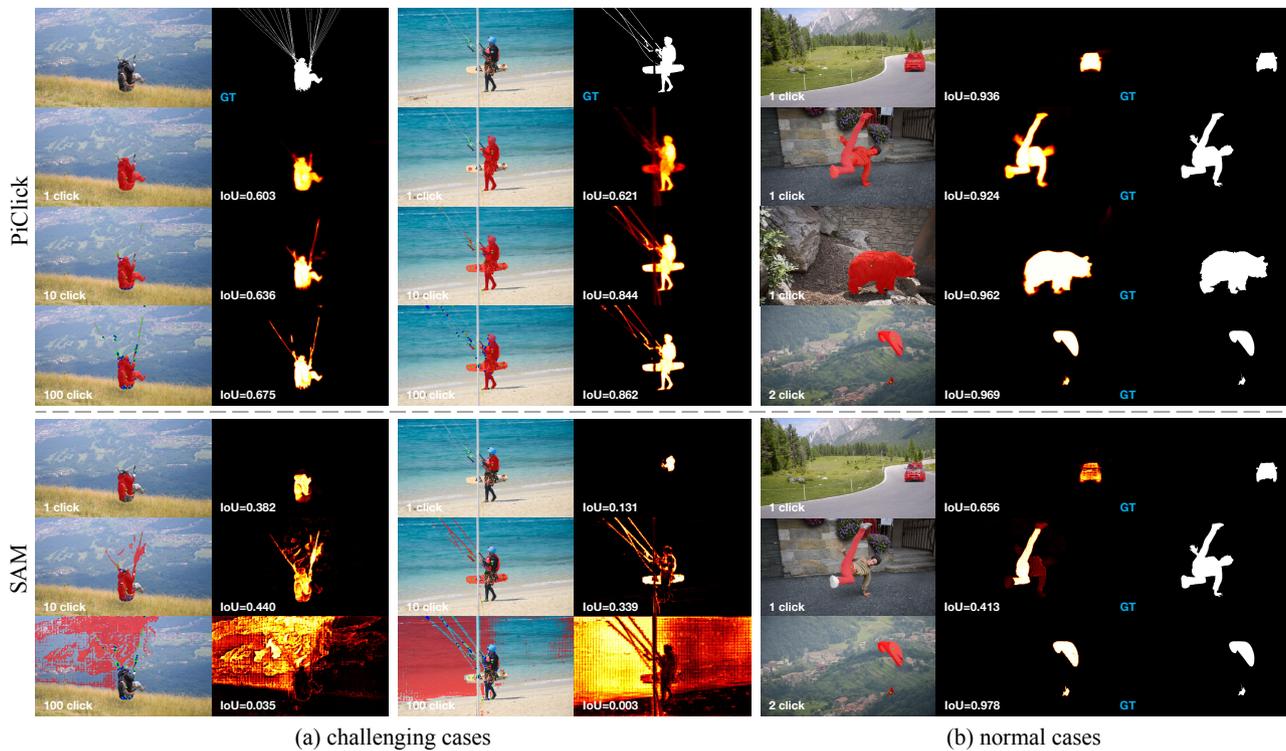

(a) challenging cases  (b) normal cases

Figure 9: Segmentation results of PiClick and SAM [26] on DAVIS [42] (a) challenging cases; (b) normal cases. The segmentation probability maps are shown in red; the segmentation maps are overlaid in red on the original images. The clicks are shown as green (positive click) or blue (negative click) dots on the image. The "GT" denotes ground truth.

With the emergence of SAM [26], we are witnessing the era of generalist segmentation models [26, 59], leveraging super large-scale datasets for training. However, we observe that SAM performs even worse in those cases. We hope PiClick will serve as a strong baseline to inspire research for better adaptation of generalist segmentation models to specific tasks.

## 5. Conclusion

In this paper, we present PiClick, a Transformer-based model to solve the target ambiguity issue in interactive segmentation. PiClick generates multiple diverse segmentation masks with a set of mask queries. A Target Reasoning module (TRM) is proposed to effectively suggest the target object from multiple proposals, further easing the user selection process. Extensive experiments show that PiClick achieves state-of-the-art performance on 6 natural image benchmarks and generalize well to medical datasets.

## References

[1] Acuna, D., Ling, H., Kar, A., Fidler, S., 2018. Efficient interactive annotation of segmentation datasets with polygon-rnn++ .

[2] Ambellan, F., Tack, A., Ehlke, M., Zachow, S., 2019. Automated segmentation of knee bone and cartilage combining statistical shape knowledge and convolutional neural networks: Data from the osteoarthritis initiative. Medical image analysis 52, 109–118.

[3] Baid, U., Ghodasara, S., Mohan, S., Bilello, M., Calabrese, E., Colak, E., Farahani, K., Kalpathy-Cramer, J., Kitamura, F.C., Pati, S., et al., 2021. The rsna-asnr-miccai brats 2021 benchmark on brain tumor segmentation and radiogenomic classification. arXiv preprint arXiv:2107.02314 .

[4] Benenson, R., Popov, S., Ferrari, V., 2019. Large-scale interactive object segmentation with human annotators , 11700–11709.

[5] Boykov, Y.Y., Jolly, M.P., 2001. Interactive graph cuts for optimal boundary & region segmentation of objects in nd images 1, 105–112.

[6] Carion, N., Massa, F., Synnaeve, G., Usunier, N., Kirillov, A., Zagoruyko, S., 2020. End-to-end object detection with transformers , 213–229.

[7] Castrejon, L., Kundu, K., Urtasun, R., Fidler, S., 2017. Annotating object instances with a polygon-rnn , 5230–5238.

[8] Chen, X., Zhao, Z., Yu, F., Zhang, Y., Duan, M., 2021. Conditional diffusion for interactive segmentation , 7345–7354.

[9] Chen, X., Zhao, Z., Zhang, Y., Duan, M., Qi, D., Zhao, H., 2022. Focalclick: towards practical interactive image segmentation , 1300–1309.

[10] Cheng, B., Misra, I., Schwing, A.G., Kirillov, A., Girdhar, R., 2022. Masked-attention mask transformer for universal image segmentation , 1290–1299.

[11] Deng, J., Dong, W., Socher, R., Li, L.J., Li, K., Fei-Fei, L., 2009. Imagenet: A large-scale image database .

[12] Ding, H., Cohen, S., Price, B., Jiang, X., 2020. Phraseclick: toward achieving flexible interactive segmentation by phrase and click , 417–435.

[13] Dosovitskiy, A., Beyer, L., Kolesnikov, A., Weissenborn, D., Zhai, X., Unterthiner, T., Dehghani, M., Minderer, M., Heigold, G., Gelly, S., et al., 2020. An image is worth 16x16 words: Transformers for image recognition at scale. arXiv preprint arXiv:2010.11929 .

[14] Dupont, C., Ouakrim, Y., Pham, Q.C., 2021. Ucp-net: unstructured contour points for instance segmentation .

[15] Everingham, M., Van Gool, L., Williams, C.K., Winn, J., Zisserman, A., 2010. The pascal visual object classes (voc) challenge. International journal of computer vision 88, 303–338.

[16] Gao, Y., Lang, C., Liu, F., Cao, Y., Sun, L., Wei, Y., 2023. Dynamic interaction dilation for interactive human parsing. IEEE Transactions on Multimedia .

[17] Gerhard, S., Funke, J., Martel, J., Cardona, A., Fetter, R., 2013. Segmented anisotropic sstem dataset of neural tissue. figshare , 0–0.

[18] Grady, L., 2006. Random walks for image segmentation. IEEE transactions on pattern analysis and machine intelligence 28.

[19] Gulshan, V., Rother, C., Criminisi, A., Blake, A., Zisserman, A., 2010. Geodesic star convexity for interactive image segmentation , 3129–3136.

[20] Gupta, A., Dollar, P., Girshick, R., 2019. Lvis: A dataset for large vocabulary instance segmentation .

[21] Hariharan, B., Arbeláez, P., Bourdev, L., Maji, S., Malik, J., 2011. Semantic contours from inverse detectors , 991–998.

[22] He, K., Chen, X., Xie, S., Li, Y., Dollár, P., Girshick, R., 2022. Masked autoencoders are scalable vision learners .

[23] Huang, Y., Yang, H., Sun, K., Zhang, S., Jiang, G., Ji, R., Cao, L., 2023. Interformer: Real-time interactive image segmentation. arXiv preprint arXiv:2304.02942 .

[24] Jain, S.D., Grauman, K., 2019. Click carving: Interactive object segmentation in images and videos with point clicks. International Journal of Computer Vision 127, 1321–1344.

[25] Kim, K., Jung, S.W., 2017. Interactive image segmentation using semi-transparent wearable glasses. IEEE Transactions on Multimedia 20, 208–223.

[26] Kirillov, A., Mintun, E., Ravi, N., Mao, H., Rolland, C., Gustafson, L., Xiao, T., Whitehead, S., Berg, A.C., Lo, W.Y., Dollár, P., Girshick, R., 2023. Segment anything arXiv:2304.02643.

[27] Le, H., Mai, L., Price, B., Cohen, S., Jin, H., Liu, F., 2018. Interactive boundary prediction for object selection , 18–33.

[28] Li, K., Tao, W., 2015. Adaptive optimal shape prior for easy interactive object segmentation. IEEE Transactions on Multimedia 17, 994–1005.

[29] Li, Z., Chen, Q., Koltun, V., 2018. Interactive image segmentation with latent diversity , 577–585.

[30] Liew, J., Wei, Y., Xiong, W., Ong, S.H., Feng, J., 2017. Regional interactive image segmentation networks , 2746–2754.

[31] Liew, J.H., Cohen, S., Price, B., Mai, L., Ong, S.H., Feng, J., 2019. Multiseg: Semantically meaningful, scale-diverse segmentations from minimal user input , 662–670.

[32] Lin, J., Chen, J., Yang, K., Roitberg, A., Li, S., Li, Z., Li, S., 2023. Adaptiveclick: Clicks-aware transformer with adaptive focal loss for interactive image segmentation. arXiv preprint arXiv:2305.04276 .

[33] Lin, T.Y., Goyal, P., Girshick, R., He, K., Dollár, P., 2017. Focal loss for dense object detection , 2980–2988.

[34] Lin, T.Y., Maire, M., Belongie, S., Hays, J., Perona, P., Ramanan, D., Dollár, P., Zitnick, C.L., 2014. Microsoft coco: Common objects in context , 740–755.

[35] Lin, Z., Duan, Z.P., Zhang, Z., Guo, C.L., Cheng, M.M., 2022. Focuscut: Diving into a focus view in interactive segmentation , 2637–2646.

[36] Lin, Z., Zhang, Z., Chen, L.Z., Cheng, M.M., Lu, S.P., 2020. Interactive image segmentation with first click attention .

[37] Liu, Q., Xu, Z., Bertasius, G., Niethammer, M., 2023. Simpleclick: Interactive image segmentation with simple vision transformers arXiv:2210.11006.

[38] Liu, Q., Zheng, M., Planche, B., Karanam, S., Chen, T., Niethammer, M., Wu, Z., 2022. Pseudoclick: Interactive image segmentation with click imitation , 728–745.

[39] Maninis, K.K., Caelles, S., Pont-Tuset, J., Van Gool, L., 2018. Deep extreme cut: From extreme points to object segmentation , 616–625.

[40] Martin, D., Fowlkes, C., Tal, D., Malik, J., 2001. A database of human segmented natural images and its application to evaluating segmentation algorithms and measuring ecological statistics 2, 416–423.

[41] Milletari, F., Navab, N., Ahmadi, S.A., 2016. V-net: Fully convolutional neural networks for volumetric medical image segmentation , 565–571.

[42] Perazzi, F., Pont-Tuset, J., McWilliams, B., Van Gool, L., Gross, M., Sorkine-Hornung, A., 2016. A benchmark dataset and evaluation methodology for video object segmentation , 724–732.

[43] Rother, C., Kolmogorov, V., Blake, A., 2004. " grabcut" interactive foreground extraction using iterated graph cuts. ACM transactions on graphics (TOG) 23, 309–314.

[44] Sakinis, T., Milletari, F., Roth, H., Korfiatis, P., Kostandy, P., Philbrick, K., Akkus, Z., Xu, Z., Xu, D., Erickson, B.J., 2019. Interactive segmentation of medical images through fully convolutional neural networks. arXiv preprint arXiv:1903.08205 .

[45] Sofiiuk, K., Petrov, I.A., Konushin, A., 2022. Reviving iterative training with mask guidance for interactive segmentation , 3141–3145.

[46] Vaswani, A., Shazeer, N., Parmar, N., Uszkoreit, J., Jones, L., Gomez, A.N., Kaiser, Ł., Polosukhin, I., 2017. Attention is all you need. Advances in neural information processing systems 30.

[47] Wang, H., Wang, S., Yan, C., Jiang, X., Tang, X., Hu, Y., Xie, W., Gavves, E., 2023a. Towards open-vocabulary video instance segmentation. arXiv preprint arXiv:2304.01715 .

[48] Wang, T., Ji, Z., Sun, Q., Chen, Q., Jing, X.Y., 2016. Interactive multilabel image segmentation via robust multilayer graph constraints. IEEE Transactions on Multimedia 18, 2358–2371.

[49] Wang, T., Ji, Z., Yang, J., Sun, Q., Fu, P., 2020. Global manifold learning for interactive image segmentation. IEEE Transactions on Multimedia 23, 3239–3249.

[50] Wang, T., Li, H., Zheng, Y., Sun, Q., 2023b. One-click-based perception for interactive image segmentation. IEEE Transactions on Neural Networks and Learning Systems .

[51] Wu, J., Zhao, Y., Zhu, J.Y., Luo, S., Tu, Z., 2014. Milcut: A sweeping line multiple instance learning paradigm for interactive image segmentation , 256–263.

[52] Xiang, S., Pan, C., Nie, F., Zhang, C., 2011. Interactive image segmentation with multiple linear reconstructions in windows. IEEE Transactions on Multimedia 13, 342–352.

[53] Xie, H., Wang, C., Zheng, M., Dong, M., You, S., Fu, C., Xu, C., 2023. Boosting semi-supervised semantic segmentation with probabilistic representations 37, 2938–2946.

[54] Xu, N., Price, B., Cohen, S., Yang, J., Huang, T., 2017. Deep grabcut for object selection. arXiv preprint arXiv:1707.00243 .

[55] Xu, N., Price, B., Cohen, S., Yang, J., Huang, T.S., 2016. Deep interactive object selection , 373–381.

[56] Zhang, S., Liew, J.H., Wei, Y., Wei, S., Zhao, Y., 2020. Interactive object segmentation with inside-outside guidance , 12234–12244.

[57] Zhang, X., Yang, K., Lin, J., Yuan, J., Li, Z., Li, S., 2023. Vpuformer: Visual prompt unified transformer for interactive image segmentation. arXiv preprint arXiv:2306.06656 .

[58] Zhou, T., Li, L., Bredell, G., Li, J., Unkelbach, J., Konukoglu, E., 2023. Volumetric memory network for interactive medical image segmentation. Medical image analysis 83, 102599.

[59] Zou, X., Yang, J., Zhang, H., Li, F., Li, L., Gao, J., Lee, Y.J., 2023. Segment everything everywhere all at once arXiv:2304.06718.